\documentclass[twoside,11pt]{article}

\usepackage{blindtext}

\usepackage[abbrvbib, preprint]{jmlr2e}

\usepackage{xcolor}
\usepackage{amsmath,amssymb}
\usepackage{lmodern}

\usepackage{longtable,booktabs,array}
\usepackage{calc} % for calculating minipage widths
\usepackage{etoolbox}

\usepackage[]{natbib}

\usepackage{booktabs}
\usepackage{longtable}
\usepackage{array}
\usepackage{multirow}
\usepackage{wrapfig}
\usepackage{float}
\usepackage{colortbl}
\usepackage{pdflscape}
\usepackage{tabu}
\usepackage{threeparttable}
\usepackage{threeparttablex}
\usepackage[normalem]{ulem}
\usepackage{makecell}
\usepackage{xcolor}
\usepackage{orcidlink}

\usepackage{lastpage}
\jmlrheading{23}{2025}{1-\pageref{LastPage}}{1/21; Revised 5/22}{9/22}{21-0000}{Leonardo Fonseca Larrubia, Pedro Alberto Morettin and Chang Chiann}

\ShortHeadings{MODWST and Its Application in Classification Tasks}{Larrubia, Morettin and Chiann}
\firstpageno{1}

\title{The Maximal Overlap Discrete Wavelet Scattering Transform and Its Application in Classification Tasks}
\author{\name  Leonardo Fonseca Larrubia \email leonardo.larrubia@usp.br \\
       University of São Paulo\\
       São Paulo, SP, Brazil
       \AND
       \name Pedro Alberto Morettin \email pam@ime.usp.br \\
       University of São Paulo\\
       São Paulo, SP, Brazil
       \AND
       \name Chang Chiann \email chang@ime.usp.br \\
       University of São Paulo\\
       São Paulo, SP, Brazil}

\editor{My editor}

\begin{document}

\maketitle

\begin{abstract}%   <- trailing '%' for backward compatibility of .sty file
We present the Maximal Overlap Discrete Wavelet Scattering Transform
(MODWST), whose construction is inspired by the combination of the
Maximal Overlap Discrete Wavelet Transform (MODWT) and the Scattering
Wavelet Transform (WST). We also discuss the use of MODWST in
classification tasks, evaluating its performance in two applications:
stationary signal classification and ECG signal classification. The
results demonstrate that MODWST achieved good performance in both
applications, positioning itself as a viable alternative to popular
methods like Convolutional Neural Networks (CNNs), particularly when the
training data set is limited.
\end{abstract}


\section{Introduction}\label{introduction}

Most classification problems involve finding data representations that
reduce variability among elements that belong to the same class while
maximizing the distance among elements belonging to different classes.
Highly correlated data, such as time series, images, and even volumetric
data, are often more challenging due to the inherent dependence between
observations and high dimensionality. Classification methods must be
robust to certain deformations, as a time series shifted in time or
subjected to slight deformations still belongs to the same class. Thus,
time series classification techniques should aim to derive
representations of the original series that are invariant to specific
input deformations.

A simple way to classify time series is to use using wavelets. This can
be done by defining a wavelet dictionary and using it to extract
features from the time series that can later serve as input for standard
classification methods. The dictionary's composition, including the type
of wavelets and the classifier used in combination, is crucial for the
method's success.

The Discrete Wavelet Transform (DWT) is an example of pattern extraction
from a time series using compactly supported wavelets that form a basis.
However, the DWT is sensitive to translations and restricted to time
series of size \(2^J\). An alternative is the Maximal Overlap Discrete
Wavelet Transform (MODWT), which is not affected by the initial offset
of the time series and can be computed for series of any size. However,
this representation is highly redundant, with its coefficients being
highly correlated, and it does not perform well in classification tasks.
Therefore, modifications and enhancements to the MODWT are needed to
improve its performance. For these enhancements, we draw inspiration
from the Wavelet Scattering Transform (WST) proposed by
\citet{Mallat2012} to construct the Maximal Overlap Discrete Wavelet
Scattering Transform (MODWST), which is a type of WST built using the
MODWT as its base.

The plan of this work is as follows. In Section \ref{sec-descriptionWST}
we give a brief description of the Wavelet Scattering Transform and in
Section \ref{sec-DWTandMODWT} we describe the Discrete and the Maximal
Overlap Discrete Wavelet Transforms. In Section \ref{sec-MODWST} we
present the Maximal Overlap Discrete Wavelet Scattering Transform. In
Section \ref{sec-applications} we give two applications and conclude in
Section \ref{sec-conclusion}.

\section{Brief Description of WST}\label{sec-descriptionWST}

The Wavelet Scattering Transform (WST) aims to construct representations
invariant to small translations and deformations, making it an excellent
tool for signal classification applications. It reduces variability
among elements within the same class while effectively distinguishing
those in different classes. Its construction is based on wavelet
functions, i.e., functions \(\psi(\cdot)\) such that
\(\int_{-\infty}^{\infty} \psi(t) dt = 0\) and
\(\int_{-\infty}^{\infty} |\psi(t)|^2 dt < \infty\).

Once an appropriate set of wavelet has been defined, WST consists of the
cascade application of the modulus function and convolutions. More
precisely, let \(f(t)\) be a signal, \(\phi\) an average filter,
i.e.~\(\int_{-\infty}^{\infty} \phi(t) dt = 1\), and
\(\left\{\psi_{\lambda_k}, \lambda_k \in \Lambda_k, k =1, ..., n \right\}\)
a sequence of wavelets functions, where the sequence
\(p = \{\lambda_1, ..., \lambda_n\}\) is called path. The WST
coefficients are obtained by

\[
\begin{aligned}
    S_k[p] f(t) &= | \ | \dots | f \star \psi_{\lambda_1}(t)| \star \dots | \star \psi_{\lambda_k}(t)| \star \phi(t),
\end{aligned}
\]

\noindent where \(\star\) indicates convolution. The WST coefficients
can be arranged in the vector
\(Sf= \left( S_k[p] f\right)_{0\leq k \leq n}.\)

Overall, the WST has a structure similar to the well-known Convolutional
Neural Networks (CNN), a technique prominent in complex signal
classification tasks. CNNs consist of sequences of convolutions and
nonlinearities applied in cascades, obtaining a final representation
that is easier to separate. In CNNs, the filter weights are learned,
leading to a vast number of parameters to estimate. In contrast, the WST
uses fixed wavelet filters. \citet{Mallat2016} discusses the functioning
of CNNs, drawing parallels with the WST structure, while
\citet{ZarkaETAL2021separation} further explores this discussion.

Several studies have applied the WST in time series analysis and signal
classification. Notable examples include the work of
\citet{AndenMallat2014}, who applied the WST to audio classification
problems, analyzing the GTZAN data set for music genre classification and
the TIMIT data set for phoneme classification;
\citet{LeonarduzziETAL2019}, who used the WST to study entropy in
financial time series; and \citet{AhmadETAL2017}, who employed the WST
to extract information from ECG data.

The two-dimensional (2D) version of the WST is also rotation-invariant,
making it suitable for image analysis. In 2D WST applications, notable
studies include those by \citet{BrunaMallat2013}, who used the WST for
image classification problems, employing it as preprocessing for
analyzing the MNIST and CIFAR-10 data sets; \citet{LaurentMallat2013},
who described the WST's properties and applied it to texture image
classification problems; \citet{OyallonMallat2015}, who analyzed the
CIFAR-10 and Caltech-101 data sets for classification, exploring wavelet
rotations; and the works by \citet{OyallonETAL2017},
\citet{OyallonETAL2019}, and \citet{zarkaETAL2020}, which created hybrid
neural network architectures where the WST serves as the first CNN
layer. Other studies include \citet{VillarMorgenshtern2021}, who used
the WST for clustering images combined with ``Projection onto Orthogonal
Complement''; \citet{ChengETAL2020}, who analyzed interstellar space
images with the WST; and \citet{ChengETAL2016}, who introduced an
orthogonal scattering transform using Haar wavelets, a deep network
computed with a hierarchy of additions, subtractions, and absolute
values over pairs of coefficients, applied to the MNIST data set for
classification.

There are also applications of the WST for 3D data analysis, as seen in
\citet{HirnETAL2017}, who introduced multiscale invariant dictionaries
based on 3D WST for quantum estimation of chemical energies of organic
molecules using training data sets; and \citet{EickenbergETAL2018}, who
presented a machine learning algorithm for predicting molecular
properties inspired by density functional theory ideas, calculating the
3D WST of estimated electronic densities of molecules.

It is worth mentioning the Kymatio software \citep{AndreuxETAL2020},
which produces the WST using Morlet wavelets. It is implemented in
Python and it is integrates with \emph{Numpy}, \emph{Keras}, and
\emph{Pytorch}.

\section{DWT and MODWT}\label{sec-DWTandMODWT}

Let \(\mathbf{X}\) be a time series of length \(T = 2^J\). The Discrete
Wavelet Transform (DWT) is an orthonormal linear transformation
expressed as\\
\begin{equation}\phantomsection\label{eq-DWT}{  
    \mathbf{W} = \mathcal{W}\mathbf{X},  
}\end{equation}

\noindent where \(\mathbf{W}\) is the vector of wavelet transform
coefficients and \(\mathcal{W}\) is the orthonormal matrix, i.e.,
\(\mathcal{W}^{\top}\mathcal{W} = I_T\), called the wavelet matrix.

The brute-force computation of equation (\ref{eq-DWT}) requires a number
of operations on the order of \(O(T^2)\). However, this transformation
can be performed more efficiently using the well-known \emph{pyramid
algorithm}, which reduces the number of calculations to just \(O(T)\).
For this purpose, consider the partitioning of the vector \(\mathbf{W}\)
and the matrix \(\mathcal{W}\) from (\ref{eq-DWT}) into, respectively,
\(J+1\) subvectors and \(J+1\) submatrices, such that

\begin{equation}\phantomsection\label{eq-DWT_SubMat}{
    \mathbf{W} = 
        \begin{bmatrix}
            \mathbf{W}_1 \\ \mathbf{W}_2 \\ \vdots \\ \mathbf{W}_J \\ \mathbf{V}_J
        \end{bmatrix} =
        \begin{bmatrix}
            \mathcal{W}_1 \\ \mathcal{W}_2 \\ \vdots \\ \mathcal{W}_J \\ \mathcal{V}_J
        \end{bmatrix}
        \mathbf{X}
    = \mathcal{W}\mathbf{X}.
}\end{equation}

Each vector \(\mathbf{W}_j\) has a size of \(T/2^j\), while each matrix
\(\mathcal{W}_j\) has dimensions \(T/2^j \times T\). To specify each of
these components, we need to introduce the wavelet filter
\(\left\{h_l \right\}\), with \(l = 0, \dots, L-1\), of size \(L\),
where \(L\) is an even integer, with \(h_0\) and \(h_{L-1}\) nonzero and
\(h_l = 0\) for \(l < 0\) and \(l \geq L\). The wavelet filter
\(\left\{h_l\right\}\) must satisfy the following conditions:

\begin{equation}\phantomsection\label{eq-DiscFiltSomaZero}{ 
    \sum_{l=0}^{L-1} h_l = 0,
}\end{equation}
\begin{equation}\phantomsection\label{eq-DiscFiltEnergia}{
    \sum_{l=0}^{L-1} h^2_l = 1,
}\end{equation}
\begin{equation}\phantomsection\label{eq-DiscFiltOrtonormal}{
    \sum_{l=0}^{L-1} h_l h_{l+2n} =  \sum_{l=-\infty}^{\infty} h_l h_{l+2n}= 0, \quad \forall n \in \mathbb{Z}.
}\end{equation}

We also need the filter \(\left\{g_l\right\}\), which is closely related
to the wavelet filter \(\left\{h_l\right\}\) through the
\emph{quadrature mirror relationship}.
\begin{equation}\phantomsection\label{eq-DiscFiltScaling}{ 
    g_l = (-1)^{l+1} h_{L-1-l},
}\end{equation}

\noindent while the inverse relationship is given by
\begin{equation}\phantomsection\label{eq-DiscFiltScalingInv}{
    h_l = (-1)^{l}g_{L-1-l}.
}\end{equation}

Letting \(\mathbf{V}_0 = \mathbf{X}\), the \(j\)-th stage of pyramid
algorithm, with \(j = 1, \dots, J\), consists of obtaining
\(\mathbf{W}_j\) and \(\mathbf{V}_j\), of sizes \(T_j = T/2^j\), such
that \[
    W_{j,t} = \sum_{l=0}^{L-1} h_l V_{j-1, 2t+1-l \mod T_{j-1}},
\] \[
    V_{j,t} = \sum_{l=0}^{L-1} g_l V_{j-1, 2t+1-l \mod T_{j-1}}.
\]

To define the Maximal Overlap Discrete Wavelet Transform (MODWT), we
first need to establish the filters \(\left\{\tilde{h}_l\right\}\) and
\(\left\{\tilde{g}_l\right\}\) required for its construction. These
filters are defined as \[ \tilde{h}_{l} = 2^{-j/2}{h}_{l},  \]
\[ \tilde{g}_{l} = 2^{-j/2}{g}_{l}. \]

Once the filters \(\left\{\tilde{h}_l\right\}\) and
\(\left\{\tilde{g}_l\right\}\) are established, we can present a
definition of the MODWT, which, similar to the DWT, consists of a
pyramid algorithm. Thus, letting
\(\widetilde{\mathbf{V}}_0 = \mathbf{X}\), the level \(j\) coefficients
of the MODWT are represented, respectively, by
\(\widetilde{\mathbf{W}}_j\) and \(\widetilde{\mathbf{V}}_j\), whose
elements are given by \[
    \widetilde{W}_{j,t}  = \sum_{l=0}^{L-1} \tilde{h}_l \tilde{V}_{j-1, t-2^{j-1}l \mod T},
\] \[
    \widetilde{V}_{j,t} = \sum_{l=0}^{L-1} \tilde{g}_l \tilde{V}_{j-1, t-2^{j-1}l \mod T}.
\]

Further description and properties of DWT and MODWT can be seen in
\citet{PercivalWalden2000}.

\section{MODWST}\label{sec-MODWST}

Before describing the construction process of the Maximal Overlap
Discrete Wavelet Scattering Transform (MODWST), we need to establish an
average filter \(\left\{\phi_l\right\}, l = 0, \dots, M\), of size
\(M < T\), such that

\[
    \sum_{l=-\infty}^{\infty}\phi_l = \sum_{l=0}^{M-1}\phi_l = 1.
\]

\noindent In addition to the average filter, it is also necessary to
define a stride value \(k\), such that \(1 \leq k \leq M\).

The first step is to apply the average filter \(\left\{\phi_l\right\}\)
to the time series \(\mathbf{X}\) of lenght \(T\), obtaining the
zeroth-order coefficients:

\[
    \widetilde{S}_{(0), i} = \sum_{l=0}^{M-1} \phi_l X_{ki+l \mod T}, \quad i=0, \cdots, \lfloor T/k \rfloor,
\]

\noindent which are assembled into the vector
\(\widetilde{\mathbf{S}}_{(0)} = \left[\widetilde{S}_{(0); 1}, \cdots, \widetilde{S}_{(0);  \lfloor T/k \rfloor} \right]\).

The second step involves selecting the discrete wavelet filter
\(\left\{h_l \right\}\), applying the MODWT to the time series
\(\mathbf{X}\) and then applying the modulus operator to each
coefficient, yielding \[  
|\widetilde{\mathbf{W}}| =  \left[ |\widetilde{\mathbf{W}}|_{(1)}, |\widetilde{\mathbf{W}}|_{(2)}, \cdots, |\widetilde{\mathbf{W}|}_{(J)} \right].
\]

After this, the firt-order coefficients of the MODWST are obtained by
the following calculation:

\[
    \widetilde{S}_{(j_1); i} = \sum_{l=0}^{M-1} \phi_l |\widetilde{W}|_{(j_1); ki+l \mod T}, \quad j_1 = 1, \dots J, \quad i=0, \cdots, \lfloor T/k \rfloor.
\]

\noindent This coefficients are assembled into the vector
\(\widetilde{\mathbf{S}}_{(j_1)} = \left[\widetilde{S}_{(j_1); 1}, \cdots, \widetilde{S}_{(j_1);  \lfloor T/k \rfloor} \right]\).

The idea of applying the modulus to the coefficients and then an average
filter of size \(M\) is to measure the amount of local energy at scale
\(i\). Thus, we can interpret the coefficients
\(\widetilde{\mathbf{S}}_{(j_1)}\) as the amount of energy at scales
\(2^{j_1}\) contained within windows of size \(M\), which slide over the
time series every \(k\) observations, i.e., each coefficient
\(\widetilde{S}_{(j_1); i}\) represents the amount of energy at scales
\(2^{j_1}\) contained by the \(i\)-th window of size \(M\).

The second-order coefficients are obtained by applying the same process
to each vector \(|\widetilde{\mathbf{W}}|_{(j)}\) generated during the
firt-order calculation process, which can be described in the following
calculation: \[
    \widetilde{S}_{(j_1, j_2); i} = \sum_{l=0}^{L-1} \phi_l |\widetilde{W}|_{(j_1 j_2); ki+l \mod T}, \quad j_1, j_2 = 1, \dots J, \quad i=0, \cdots, \lfloor T/k \rfloor.
\]

Note that for each level \(j_1 = 1, \dots, J\) of depth 1, we obtain the
vectors \(\widetilde{\mathbf{W}}_{(j_1, j_2)}\), with
\(j_2 = 1, \dots, J\), which depend on the respective first-order
coefficients. In other words, \(\widetilde{\mathbf{W}}_{(j_1, j_2)}\)
form a hierarchical dependency with the level \(j_1\) of depth 1.

We can generalize the calculation for any maximum depth \(m\). To do so,
we will use the concept of a level path, which is defined by the
sequence \(p_n = (j_1, \cdots, j_n)\), representing a level path of
maximum depth \(n\). Therefore, given an order path
\(p_n = (j_1, \cdots, j_n)\), an average filter
\(\left\{ \phi_l \right\}\) of size \(M\), a stride value \(k \leq M\),
a wavelet filter \(\left\{ h_l \right\}\) and its respective scaling
filter \(\left\{ g_l \right\}\), both of size \(L\), the MODWST
coefficients of depth \(m\), with \(1 \leq m \leq n\), are given by \[
    \widetilde{S}_{p_m; i} = \sum_{l=0}^{M-1} \phi_l |\widetilde{W}|_{(j_1, \cdots, j_m); ki+l \mod T}, \quad i=0, \cdots, \lfloor T/k \rfloor,
\]

\noindent where \(|\widetilde{W}|_{(j_1, \cdots, j_{m}); t}\) is the
modulus of the coefficients \(\widetilde{W}_{(j_1, \cdots, j_{m}); t}\),
obtained by the following expressions: \[
    \widetilde{W}_{p_m; t} = \widetilde{W}_{(j_1, \cdots, j_m); t} = \sum_{l=0}^{L-1} \tilde{h}_l \widetilde{V}_{(j_1, \cdots, j_{m}-1), t-2^{j-1}l \mod T},
\] \[
    \widetilde{V}_{p_m; t} = \widetilde{V}_{(j_1, \cdots, j_m); t} = \sum_{l=0}^{L-1} \tilde{g}_l \widetilde{V}_{(j_1, \cdots, j_m-1), t-2^{j-1}l \mod T},
\] \noindent where \[
\widetilde{V}_{(j_1, \cdots, j_{m-1}, 0);t} = |\widetilde{W}|_{(j_1, \cdots, j_{m-1}); t}, \quad  0<j_i<J.
\]

\subsection{Comments}\label{sec-discutions}

The MODWST, with a maximum depth \(m\) and stride value \(k\), applied
to a time series of size \(T\), will contain at most
\(\lceil T/k \rceil \left(\sum_{i=1}^{m} \lfloor \log_2 T \rfloor^i + 1 \right)\)
coefficients. The zeroth depth is composed of \(\lceil T/k \rceil\)
coefficients, while the \(i\)-th depth is composed of
\(\lceil T/k \rceil \lfloor \log_2 T \rfloor^i\) coefficients, with
\(i > 0\).

The MODWT has a computational cost on the order of \(O(T \log T)\).
Therefore, for an MODWST with a maximum depth \(m\), its computational
cost will be on the order of \(O(T (\log T)^m)\), due to its cascading
calculations.

A remarkable feature is that the MODWST and WST share similar
properties, due to their structures being analogous. Thus, one important
property of the WST which can be extended to the MODWST is that the
deeper the MODWST, the smaller the energy of the coefficients
corresponding to those depths. This can be state in the following
result:

\[
    \left|  \widetilde{\mathbf{S}}_{(j_1,\cdots,j_m)}\right| - \left|  \widetilde{\mathbf{S}}_{(j_1, \cdots, j_m, j_{m+1})}\right| \geq 0,
\]

\noindent for all deph \(m = 1,2, \dots\) and
\(j_{m+1} = 1, \cdots, J\). This result can also be seen as a
consequence of MODWT being an energy-preserving transformation of the
original series \citep{PercivalWalden2000}. Since MODWST does not use
\(V_J\) in the cascading process, it implies that the sum of the
energies of the deeper orders will always be smaller than the energy of
the coefficients of the previous order.

\begin{figure}
  \centering{
    \includegraphics[width=0.9\textwidth]{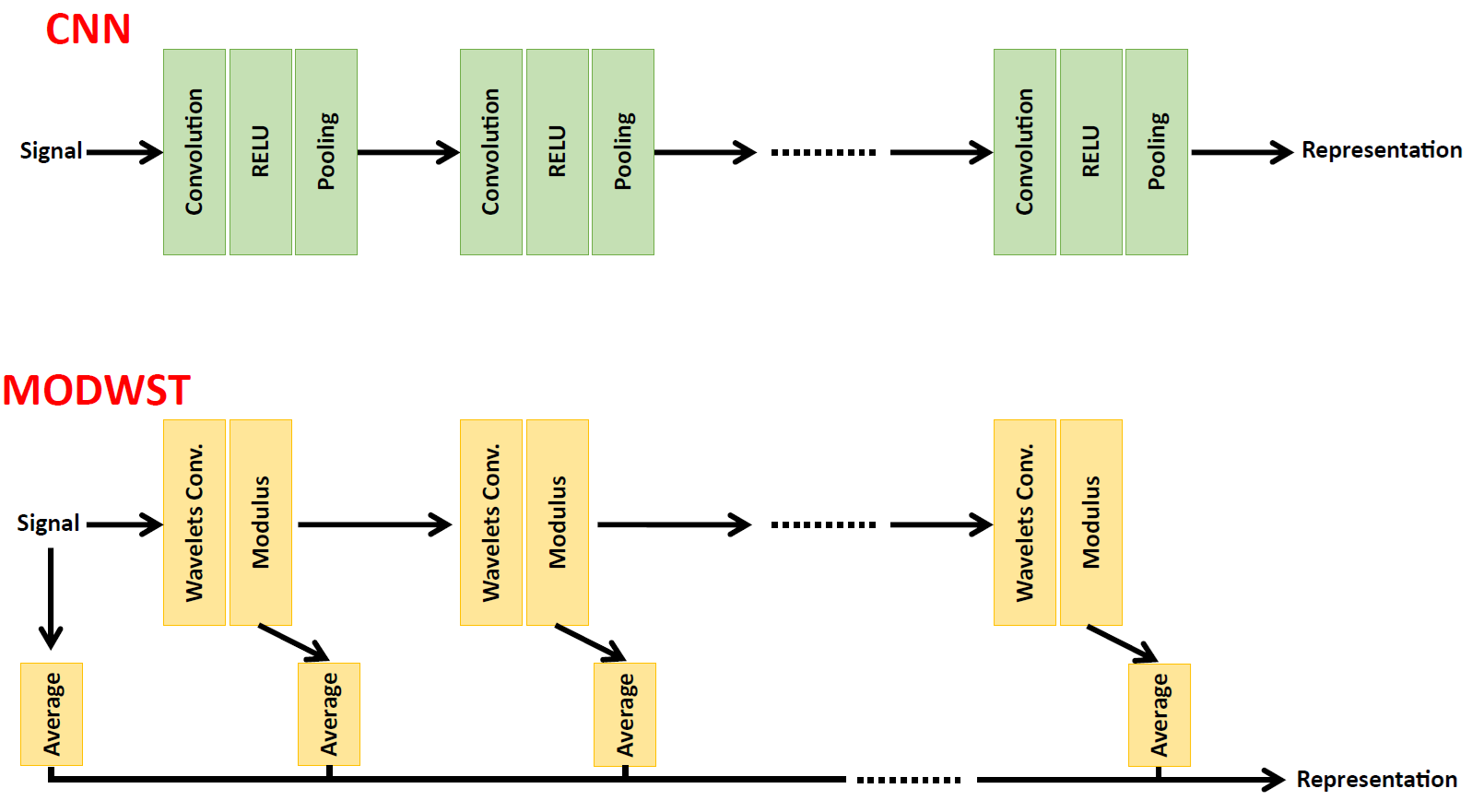}
  }
  \caption{\label{fig-SWTeCNN}Comparison of the General Structures of CNNs and MODWST.}
\end{figure}

A final feature that deserves to be highlighted is the similarity
between the MODWST and a convolutional neural network (CNN). As
highlighted by \citet{Mallat2016} regarding the WST, we can observe that
the wavelet filters \(\{h_l\}\) resemble the convolution filters learned
by CNNs, the modulus step resembles the non-linear activation functions
in CNNs, and the averaging filter \(\{g_l\}\) corresponds to the pooling
step in CNNs. Thus, the MODWST behaves analogously to CNNs but with the
computational advantage of having fixed convolution weights and the
disadvantage of lower flexibility. Figure \ref{fig-SWTeCNN} illustrates
the comparison of the structure of MODWST and CNN.

\section{Applications}\label{sec-applications}

We present in this section two applications aimed at evaluating the
performance of MODWST in classification tasks. For these applications,
we used the R software \citep{RSoftaware}, with DWT and MODWT
calculations performed using the \texttt{wavelism} \citep{waveslim}
package, and classification methods were performed using the
\texttt{e1071} \citep{e1071} and \texttt{kernlab} \citep{kernlab}
packages. Additionally, we utilized the auxiliary packages
\texttt{doParallel} \citep{doParallel} and \texttt{tidyverse}
\citep{tidyverse}. The data and scripts for the applications are
available at \url{https://github.com/laftos/MODWST}.

\subsection{Stationary Processes}\label{sec-aplStationaryProcess}

\citet{AndenMallat2014} assert that the WST is effective for
differentiating stationary processes. The goal of this application is to
verify whether the MODWST can also perform well in tasks involving the
discrimination of stationary signals. To evaluate this, we simulated
\(200\) time series of size \(2^{10}\) for each type of stationary
process. The stationary processes used in this experiment were:

\begin{itemize}
% \tightlist
\item
  \textbf{AR1a}: AR(1) process with \(\phi = 0.8\) and Normal errors
  \((0, 0.5^2)\);
\item
  \textbf{AR1b}: AR(1) process with \(\phi = 0.4\) and Normal errors
  \((0, 0.5^2)\);
\item
  \textbf{AR1c}: AR(1) process with \(\phi = -0.4\) and Normal errors
  \((0, 0.5^2)\);
\item
  \textbf{AR2}: AR(2) process with \(\phi_1 = 0.8\), \(\phi_2 = -0.9\),
  and Normal errors \((0, 0.5^2)\);
\item
  \textbf{ARMA1a}: ARMA(1,1) process with \(\phi_1 = 0.8\),
  \(\theta_1 = 0.8\), and Normal errors \((0, 0.5^2)\);
\item
  \textbf{ARMA1b}: ARMA(1,1) process with \(\phi_1 = 0.8\),
  \(\theta_1 = -0.8\), and Normal errors \((0, 0.5^2)\);
\item
  \textbf{N1}: White noise process with Normal distribution
  \((0, 0.5^2)\);
\item
  \textbf{N2}: White noise process with Normal distribution
  \((0, 2.5^2)\);
\item
  \textbf{T1}: White noise process with Student's \(t\)-distribution
  \((\nu = 1)\);
\item
  \textbf{T3}: White noise process with Student's \(t\)-distribution
  \((\nu = 3)\);
\item
  \textbf{C}: White noise process with Cauchy distribution
  \((\gamma = 0.5)\);
\item
  \textbf{U}: White noise process with Uniform distribution
  \((-0.75, 0.75)\);
\item
  \textbf{E}: White noise process with Exponential distribution
  \((\lambda = 2)\), shifted to have zero mean;
\item
  \textbf{B}: White noise Bernoulli process \((p = 0.5)\), shifted to
  have zero mean;
\item
  \textbf{P}: White noise Poisson process \((\lambda = 0.5^2)\), shifted
  to have zero mean.
\end{itemize}

\begin{figure}
  \centering{
    \includegraphics[width=0.9\textwidth]{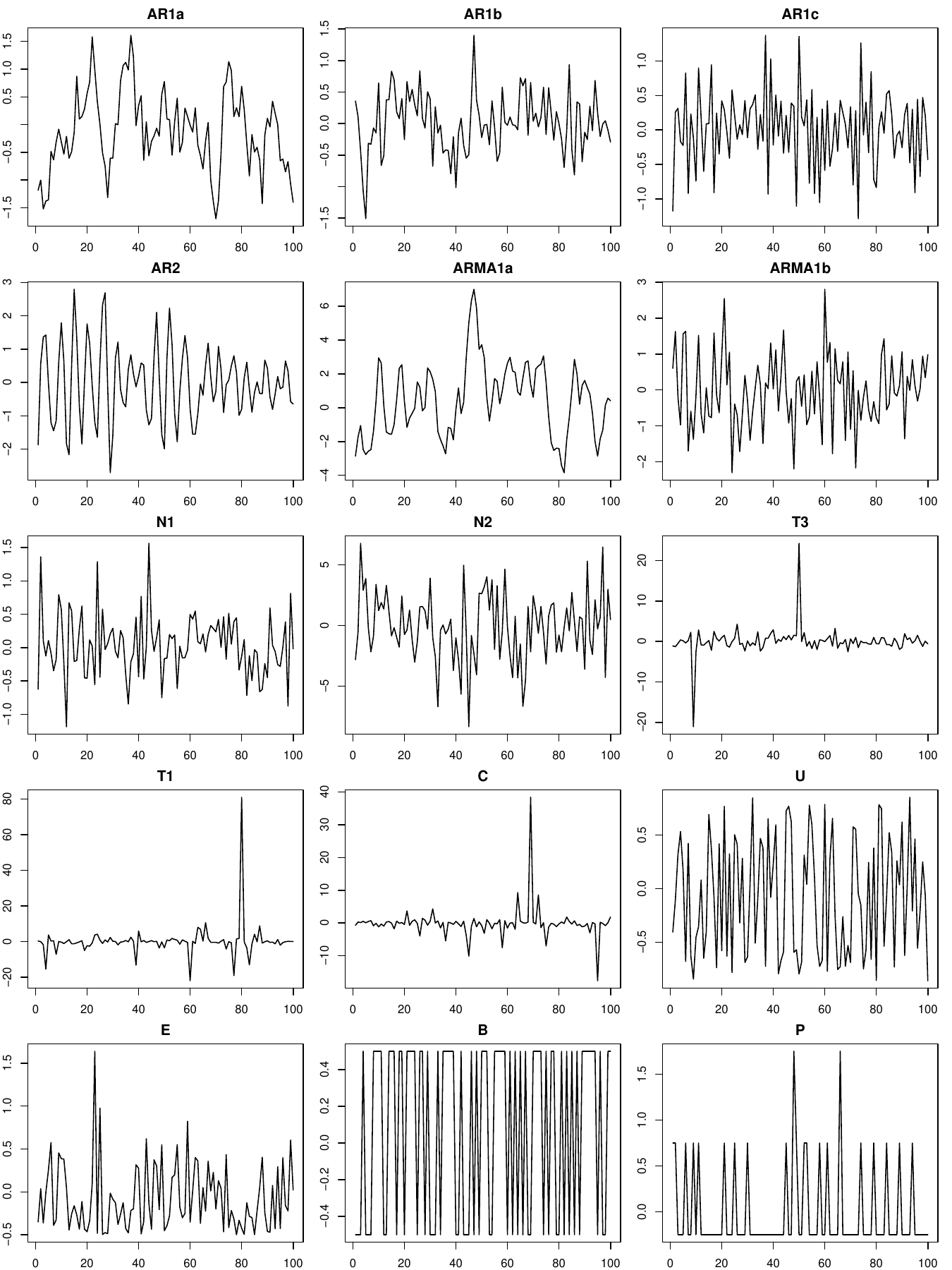}
    }
    \caption{\label{fig-SeriesSimuladas}Example of simulated series for each stationary process.}
\end{figure}

\noindent Figure \ref{fig-SeriesSimuladas} shows an example of the first
100 observations from each type of simulated series.

The data set was split into \(80\%\) training and \(20\%\) testing and
was it was ensured an equal number of observations per class in both
sets. Thus, for each class, \(160\) series were used for training and
\(40\) for testing.

We used five representations of the time series:

\begin{enumerate}
\def\labelenumi{\arabic{enumi}.}
% \tightlist
\item
  \textbf{Original}: The original time series values.
\item
  \textbf{DWT}: The DWT coefficients, using all \(2^{10}\) coefficients
  obtained.
\item
  \textbf{MODWT}: The MODWT coefficients, using all \(11 \times 2^{10}\)
  coefficients obtained.
\item
  \textbf{MODWST1}: Coefficients obtained from applying MODWST of depth
  \(1\), with a uniform averaging filter \(\{\phi_l\}\) of size \(2^5\)
  and spacing \(k = 2^4\), resulting in \(704\) coefficients.
\item
  \textbf{MODWST2}: Coefficients obtained from applying MODWST of depth
  \(2\), with the same averaging filter and spacing, resulting in
  \(7104\) coefficients. However, as coefficients for all observations
  in the path \((10, 10)\) were computationally zero, they were
  excluded, leaving \(7040\) coefficients.
\end{enumerate}

The Haar wavelet was used in all wavelet-based representations. These
representations were combined with four classification techniques:
Support Vector Machine (SVM) with linear kernel, SVM with Gaussian
kernel, Linear Discriminant Analysis (LDA) and Naive Bayes.

Table \ref{tbl-ResultadoGeralSim} summarizes the results. we can see
that MODWST2 representation (up to depth \(2\)) combined with the SVM
classifier with a linear kernel achieved the highest accuracy.

\begin{table}

\centering{

\centering
\begin{tabular}{lrrrr}
\toprule
\textbf{Classifier} & \textbf{Accuracy (\%)} & \textbf{LL (\%)} & \textbf{UL (\%)} & \textbf{Kappa}\\
\midrule
\addlinespace[0.3em]
\multicolumn{5}{l}{\textbf{Original}}\\
\hspace{1em}SVM-Linear & 17.3 & 14.39 & 20.6 & 0.1143\\
\hspace{1em}SVM-Radial & 57.8 & 53.77 & 61.8 & 0.5482\\
\hspace{1em}LDA & 18.2 & 15.16 & 21.5 & 0.1232\\
\hspace{1em}Naive Bayes & 63.0 & 59.00 & 66.9 & 0.6036\\
\addlinespace[0.3em]
\multicolumn{5}{l}{\textbf{DWT}}\\
\hspace{1em}SVM-Linear & 11.5 & 9.06 & 14.3 & 0.0518\\
\hspace{1em}SVM-Radial & 47.0 & 42.95 & 51.1 & 0.4321\\
\hspace{1em}LDA & 11.8 & 9.36 & 14.7 & \vphantom{1} 0.0554\\
\hspace{1em}Naive Bayes & 68.0 & 64.10 & 71.7 & 0.6571\\
\addlinespace[0.3em]
\multicolumn{5}{l}{\textbf{MODWT}}\\
\hspace{1em}SVM-Linear & 11.7 & 9.21 & 14.5 & 0.0536\\
\hspace{1em}SVM-Radial & 49.8 & 45.76 & 53.9 & 0.4625\\
\hspace{1em}LDA & 11.8 & 9.36 & 14.7 & 0.0554\\
\hspace{1em}Naive Bayes & 64.0 & 60.01 & 67.8 & 0.6143\\
\addlinespace[0.3em]
\multicolumn{5}{l}{\textbf{MODWST1}}\\
\hspace{1em}SVM-Linear & 83.2 & 79.93 & 86.1 & 0.8196\\
\hspace{1em}SVM-Radial & 70.7 & 66.84 & 74.3 & 0.6857\\
\hspace{1em}LDA & 79.3 & 75.87 & 82.5 & 0.7786\\
\hspace{1em}Naive Bayes & 85.0 & 81.89 & 87.8 & 0.8393\\
\addlinespace[0.3em]
\multicolumn{5}{l}{\textbf{MODWST2}}\\
\hspace{1em}SVM-Linear & 97.3 & 95.71 & 98.5 & 0.9714\\
\hspace{1em}SVM-Radial & 78.3 & 74.82 & 81.6 & 0.7679\\
\hspace{1em}LDA & 92.2 & 89.72 & 94.2 & 0.9161\\
\hspace{1em}Naive Bayes & 90.8 & 88.24 & 93.0 & 0.9018\\
\bottomrule
\end{tabular}

}
\caption{\label{tbl-ResultadoGeralSim}General results of classification
performance for various methods on simulated stationary series. The
lower limit (LL) and upper limit (UL) of the 95\% confidence interval
are reported.}

\end{table}%

Table \ref{tbl-ResSVMLinMODWST} presents the classification results for
the procedure that employed the MODWST with a maximum depth of 2 in
combination with the SVM classifier with a linear kernel. It can be
observed that the most frequent classification errors occurred only
between the ``Uniform'' and ``Normal'' classes, as well as between the
``Cauchy'' and ``t-Student(1)'' classes.

\begin{table}

\centering{

\centering
\begin{tabular}{l>{\raggedleft\arraybackslash}p{0.2cm}>{\raggedleft\arraybackslash}p{0.2cm}>{\raggedleft\arraybackslash}p{0.2cm}>{\raggedleft\arraybackslash}p{0.2cm}>{\raggedleft\arraybackslash}p{0.2cm}>{\raggedleft\arraybackslash}p{0.2cm}>{\raggedleft\arraybackslash}p{0.2cm}>{\raggedleft\arraybackslash}p{0.2cm}>{\raggedleft\arraybackslash}p{0.2cm}>{\raggedleft\arraybackslash}p{0.2cm}>{\raggedleft\arraybackslash}p{0.2cm}>{\raggedleft\arraybackslash}p{0.2cm}>{\raggedleft\arraybackslash}p{0.2cm}>{\raggedleft\arraybackslash}p{0.2cm}>{\raggedleft\arraybackslash}p{0.2cm}}
\toprule
\multicolumn{1}{c}{\textbf{ }} & \multicolumn{15}{c}{\textbf{Reference}} \\
\cmidrule(l{3pt}r{3pt}){2-16}
\rotatebox{90}{\textbf{ }} & \rotatebox{90}{\textbf{AR1a}} & \rotatebox{90}{\textbf{AR1b}} & \rotatebox{90}{\textbf{AR1c}} & \rotatebox{90}{\textbf{AR2}} & \rotatebox{90}{\textbf{ARMA1a}} & \rotatebox{90}{\textbf{ARMA1b}} & \rotatebox{90}{\textbf{B}} & \rotatebox{90}{\textbf{C}} & \rotatebox{90}{\textbf{E}} & \rotatebox{90}{\textbf{N1}} & \rotatebox{90}{\textbf{N2}} & \rotatebox{90}{\textbf{P}} & \rotatebox{90}{\textbf{T1}} & \rotatebox{90}{\textbf{T3}} & \rotatebox{90}{\textbf{U}}\\
\midrule
\addlinespace[0.3em]
\multicolumn{16}{l}{\textbf{Prediction}}\\
\hspace{1em}AR1a & 40 & 0 & 0 & 0 & 0 & 0 & 0 & 0 & 0 & 0 & 0 & 0 & 0 & 0 & 0\\
\hspace{1em}AR1b & 0 & 40 & 0 & 0 & 0 & 0 & 0 & 0 & 0 & 0 & 0 & 0 & 0 & 0 & 0\\
\hspace{1em}AR1c & 0 & 0 & 40 & 0 & 0 & 0 & 0 & 0 & 0 & 0 & 0 & 0 & 0 & 0 & 0\\
\hspace{1em}AR2 & 0 & 0 & 0 & 40 & 0 & 0 & 0 & 0 & 0 & 0 & 0 & 0 & 0 & 0 & 0\\
\hspace{1em}ARMA1a & 0 & 0 & 0 & 0 & 40 & 0 & 0 & 0 & 0 & 0 & 0 & 0 & 0 & 0 & 0\\
\hspace{1em}ARMA1b & 0 & 0 & 0 & 0 & 0 & 40 & 0 & 0 & 0 & 0 & 0 & 0 & 0 & 0 & 0\\
\hspace{1em}B & 0 & 0 & 0 & 0 & 0 & 0 & 39 & 0 & 0 & 0 & 0 & 0 & 0 & 0 & 0\\
\hspace{1em}C & 0 & 0 & 0 & 0 & 0 & 0 & 0 & 37 & 0 & 0 & 0 & 0 & 3 & 0 & 0\\
\hspace{1em}E & 0 & 0 & 0 & 0 & 0 & 0 & 0 & 0 & 40 & 0 & 0 & 0 & 0 & 0 & 0\\
\hspace{1em}N1 & 0 & 0 & 0 & 0 & 0 & 0 & 0 & 0 & 0 & 35 & 0 & 0 & 0 & 0 & 4\\
\hspace{1em}N2 & 0 & 0 & 0 & 0 & 0 & 0 & 0 & 1 & 0 & 0 & 40 & 0 & 0 & 0 & 0\\
\hspace{1em}P & 0 & 0 & 0 & 0 & 0 & 0 & 0 & 0 & 0 & 0 & 0 & 40 & 0 & 0 & 0\\
\hspace{1em}T1 & 0 & 0 & 0 & 0 & 0 & 0 & 0 & 2 & 0 & 0 & 0 & 0 & 37 & 0 & 0\\
\hspace{1em}T3 & 0 & 0 & 0 & 0 & 0 & 0 & 0 & 0 & 0 & 0 & 0 & 0 & 0 & 40 & 0\\
\hspace{1em}U & 0 & 0 & 0 & 0 & 0 & 0 & 1 & 0 & 0 & 5 & 0 & 0 & 0 & 0 & 36\\
\bottomrule
\end{tabular}

}
\caption{\label{tbl-ResSVMLinMODWST}Confusion matrix of the
classification procedure using MODWST with a maximum depth of 2 in
combination with the SVM classifier with a linear kernel.}

\end{table}%

\subsection{ECG Time Series}\label{ecg-time-series}

We present an application of the MODWST on ECG signals using the
\emph{PTB Diagnostics} electrocardiogram database, previously processed
by \citet{KachueeETAL2018}. The goal is to classify heartbeat signals to
identify potential cardiac arrhythmias.

The data set consists of 14,552 time series, which 4,046 (27.8\%) are
from patients with normal heartbeats, and 10,506 (72.2\%) are from
patients with abnormal heartbeats. The original time series have varying
lengths. To ensure uniformity, the latest values
\hspace{0pt}\hspace{0pt}of the time series was padded with zeros to
reach a total of 187 observations. The data set was split into 80\% for
training and 20\% for testing.

In this study we used the same representations of the time series
described in subsection \ref{sec-aplStationaryProcess}, except for MODWT
representation because it has a high number of coefficients and thus
excessively increases the computational cost in the classification
process. We also consider the same classification methods described in
subsection \ref{sec-aplStationaryProcess}.

The summary of the results is available in Table
\ref{tbl-ResultadoGeralECG2}, from which we can see that the best
classification method, because it presented the highest accuracy, was
the MODWST2 representation, with coefficients of up to depth 2, in
combination with the SVM classifier with linear kernel, having an
accuracy of \(98.1\%\). It is worth mentioning, for comparative
purposes, that \citet{KachueeETAL2018} proposed a CNN-based classifier
that had an accuracy of \(95.9\%\).

\begin{table}

\centering{

\centering
\begin{tabular}{lrrrr}
\toprule
\textbf{Classifier} & \textbf{Accuracy (\%)} & \textbf{LL (\%)} & \textbf{UL (\%)} & \textbf{Kappa}\\
\midrule
\addlinespace[0.3em]
\multicolumn{5}{l}{\textbf{Original}}\\
\hspace{1em}SVM-Linear & 83.4 & 82.0 & 84.7 & \vphantom{1} 0.555\\
\hspace{1em}SVM-Radial & 92.1 & 91.1 & 93.1 & \vphantom{1} 0.800\\
\hspace{1em}LDA & 77.7 & 76.1 & 79.2 & \vphantom{1} 0.510\\
\hspace{1em}Naive Bayes & 61.0 & 59.2 & 62.8 & 0.258\\
\addlinespace[0.3em]
\multicolumn{5}{l}{\textbf{DWT}}\\
\hspace{1em}SVM-Linear & 83.4 & 82.0 & 84.7 & 0.555\\
\hspace{1em}SVM-Radial & 92.1 & 91.1 & 93.1 & 0.800\\
\hspace{1em}LDA & 77.7 & 76.1 & 79.2 & 0.510\\
\hspace{1em}Naive Bayes & 73.8 & 72.2 & 75.4 & 0.458\\
\addlinespace[0.3em]
\multicolumn{5}{l}{\textbf{MODWST1}}\\
\hspace{1em}SVM-Linear & 92.2 & 91.2 & 93.2 & 0.804\\
\hspace{1em}SVM-Radial & 96.0 & 95.3 & 96.7 & 0.900\\
\hspace{1em}LDA & 87.4 & 86.1 & 88.5 & 0.704\\
\hspace{1em}Naive Bayes & 66.4 & 64.6 & 68.1 & 0.342\\
\addlinespace[0.3em]
\multicolumn{5}{l}{\textbf{MODWST2}}\\
\hspace{1em}SVM-Linear & 98.1 & 97.5 & 98.5 & 0.952\\
\hspace{1em}SVM-Radial & 97.5 & 96.9 & 98.1 & 0.938\\
\hspace{1em}LDA & 96.4 & 95.7 & 97.1 & 0.912\\
\hspace{1em}Naive Bayes & 64.3 & 62.5 & 66.1 & 0.319\\
\bottomrule
\end{tabular}

}
\caption{\label{tbl-ResultadoGeralECG2}General results of classification
performance for various methods applied to simulated stationary time
series. The lower limit (LL) and upper limit (UL) of the 95\% confidence
interval are reported.}

\end{table}%

Table \ref{tbl-ResSVMLinSMODWT_ECG2} shows the classification results of
the procedure using MODWST with a maximum depth of 2 in combination with
the SVM classifier with a linear kernel. It is observed that the types
of errors were relatively balanced between the classes.

\begin{table}

\centering{

\centering
\begin{tabular}{lrr}
\toprule
\multicolumn{1}{c}{\textbf{ }} & \multicolumn{2}{c}{\textbf{Reference}} \\
\cmidrule(l{3pt}r{3pt}){2-3}
\rotatebox{0}{\textbf{ }} & \rotatebox{0}{\textbf{0}} & \rotatebox{0}{\textbf{1}}\\
\midrule
\addlinespace[0.3em]
\multicolumn{3}{l}{\textbf{Prediction}}\\
\hspace{1em}0 & 773 (26,6\%) & 33 ( 1,1\%)\\
\hspace{1em}1 & 23 ( 0,8\%) & 2082 (71,5\%)\\
\bottomrule
\end{tabular}

}
\caption{\label{tbl-ResSVMLinSMODWT_ECG2}Confusion matrix of the
classification procedure using MODWST with a maximum depth of 2 combined
with the SVM classifier with a linear kernel. Classification of ECG data
for the diagnosis of cardiac arrhythmia.}

\end{table}%

\section{Conclusion}\label{sec-conclusion}

We presented the construction of the MODWST and discussed its use in
classification tasks. This representation achieves strong performance in
classification problems, serving as an alternative to popular methods
like CNNs, especially in scenarios with limited training observations.

Further works shall focus on a deeper study of the properties of the
MODWST and an evaluation of its performance in combination with other
classification methods (such as neural networks and random forests) as
well as in unsupervised problems, like data clustering. Additionally,
there is potential to extend the MODWST to two and three dimensional
cases, enabling its application to image classification and volumetric
data problems.

\section*{Acknowledgments}\label{acknowledgments}
\addcontentsline{toc}{section}{Acknowledgments}

This research was supported by CAPES (Leonardo Fonseca Larrubia) and
partial support of a FAPESP grant 2023/02538-0 (Pedro Alberto Morettin
and Chang Chiann).

\renewcommand\refname{References}
\bibliography{bibliography.bib}

\begin{thebibliography}{26}
\providecommand{\natexlab}[1]{#1}
\providecommand{\url}[1]{\texttt{#1}}
\expandafter\ifx\csname urlstyle\endcsname\relax
  \providecommand{\doi}[1]{doi: #1}\else
  \providecommand{\doi}{doi: \begingroup \urlstyle{rm}\Url}\fi

\bibitem[Ahmad et~al.(2017)Ahmad, Kamboh, Saleem, and Khan]{AhmadETAL2017}
M.~Z. Ahmad, A.~M. Kamboh, S.~Saleem, and A.~A. Khan.
\newblock Mallat's scattering transform based anomaly sensing for detection of
  seizures in scalp eeg.
\newblock \emph{IEEE Access}, 5:\penalty0 16919--16929, 2017.
\newblock \doi{10.1109/ACCESS.2017.2736014}.

\bibitem[Andreux et~al.(2020)Andreux, Angles, Exarchakis, Leonarduzzi,
  Rochette, Thiry, Zarka, Mallat, Andén, Belilovsky, Bruna, Lostanlen,
  Chaudhary, Hirn, Oyallon, Zhang, Cella, and Eickenberg]{AndreuxETAL2020}
M.~Andreux, T.~Angles, G.~Exarchakis, R.~Leonarduzzi, G.~Rochette, L.~Thiry,
  J.~Zarka, S.~Mallat, J.~Andén, E.~Belilovsky, J.~Bruna, V.~Lostanlen,
  M.~Chaudhary, M.~J. Hirn, E.~Oyallon, S.~Zhang, C.~Cella, and M.~Eickenberg.
\newblock Kymatio: Scattering transforms in python.
\newblock \emph{Journal of Machine Learning Research}, 21\penalty0
  (60):\penalty0 1--6, 2020.
\newblock URL \url{http://jmlr.org/papers/v21/19-047.html}.

\bibitem[Andén and Mallat(2014)]{AndenMallat2014}
J.~Andén and S.~Mallat.
\newblock Deep scattering spectrum.
\newblock \emph{IEEE Transactions on Signal Processing}, 62\penalty0
  (16):\penalty0 4114--4128, 2014.
\newblock \doi{10.1109/TSP.2014.2326991}.

\bibitem[Bruna and Mallat(2013)]{BrunaMallat2013}
J.~Bruna and S.~Mallat.
\newblock Invariant scattering convolution networks.
\newblock \emph{IEEE Transactions on Pattern Analysis \&amp; Machine
  Intelligence}, 35\penalty0 (08):\penalty0 1872--1886, aug 2013.
\newblock ISSN 1939-3539.
\newblock \doi{10.1109/TPAMI.2012.230}.

\bibitem[Cheng et~al.(2020)Cheng, Ting, Ménard, and Bruna]{ChengETAL2020}
S.~Cheng, Y.-S. Ting, B.~Ménard, and J.~Bruna.
\newblock {A new approach to observational cosmology using the scattering
  transform}.
\newblock \emph{Monthly Notices of the Royal Astronomical Society},
  499\penalty0 (4):\penalty0 5902--5914, 10 2020.
\newblock ISSN 0035-8711.
\newblock \doi{10.1093/mnras/staa3165}.
\newblock URL \url{https://doi.org/10.1093/mnras/staa3165}.

\bibitem[Cheng et~al.(2016)Cheng, Chen, and Mallat]{ChengETAL2016}
X.~Cheng, X.~Chen, and S.~Mallat.
\newblock Deep haar scattering networks.
\newblock \emph{Information and Inference: A Journal of the IMA}, 5\penalty0
  (2):\penalty0 105--133, 2016.
\newblock \doi{10.1093/imaiai/iaw007}.

\bibitem[Corporation and Weston(2022)]{doParallel}
M.~Corporation and S.~Weston.
\newblock \emph{doParallel: Foreach Parallel Adaptor for the 'parallel'
  Package}, 2022.
\newblock URL \url{https://CRAN.R-project.org/package=doParallel}.
\newblock R package version 1.0.17.

\bibitem[Eickenberg et~al.(2018)Eickenberg, Exarchakis, Hirn, Mallat, and
  Thiry]{EickenbergETAL2018}
M.~Eickenberg, G.~Exarchakis, M.~Hirn, S.~Mallat, and L.~Thiry.
\newblock {Solid harmonic wavelet scattering for predictions of molecule
  properties}.
\newblock \emph{The Journal of Chemical Physics}, 148\penalty0 (24):\penalty0
  241732, 05 2018.
\newblock ISSN 0021-9606.
\newblock \doi{10.1063/1.5023798}.
\newblock URL \url{https://doi.org/10.1063/1.5023798}.

\bibitem[Hirn et~al.(2017)Hirn, Mallat, and Poilvert]{HirnETAL2017}
M.~Hirn, S.~Mallat, and N.~Poilvert.
\newblock Wavelet scattering regression of quantum chemical energies.
\newblock \emph{Multiscale Modeling \& Simulation}, 15\penalty0 (2):\penalty0
  827--863, 2017.
\newblock \doi{10.1137/16M1075454}.
\newblock URL \url{https://doi.org/10.1137/16M1075454}.

\bibitem[Kachuee et~al.(2018)Kachuee, Fazeli, and Sarrafzadeh]{KachueeETAL2018}
M.~Kachuee, S.~Fazeli, and M.~Sarrafzadeh.
\newblock {ECG} heartbeat classification: {A} deep transferable representation.
\newblock \emph{CoRR}, abs/1805.00794, 2018.
\newblock URL \url{http://arxiv.org/abs/1805.00794}.

\bibitem[Karatzoglou et~al.(2004)Karatzoglou, Smola, Hornik, and
  Zeileis]{kernlab}
A.~Karatzoglou, A.~Smola, K.~Hornik, and A.~Zeileis.
\newblock kernlab -- an {S4} package for kernel methods in {R}.
\newblock \emph{Journal of Statistical Software}, 11\penalty0 (9):\penalty0
  1--20, 2004.
\newblock \doi{10.18637/jss.v011.i09}.

\bibitem[Leonarduzzi et~al.(2019)Leonarduzzi, Rochette, Bouchaud, and
  Mallat]{LeonarduzziETAL2019}
R.~Leonarduzzi, G.~Rochette, J.-P. Bouchaud, and S.~Mallat.
\newblock Maximum-entropy scattering models for financial time series.
\newblock In \emph{ICASSP 2019 - 2019 IEEE International Conference on
  Acoustics, Speech and Signal Processing (ICASSP)}, pages 5496--5500, 2019.
\newblock \doi{10.1109/ICASSP.2019.8683734}.

\bibitem[Mallat()]{Mallat2012}
S.~Mallat.
\newblock Group invariant scattering.
\newblock \emph{Communications on Pure and Applied Mathematics}, 65\penalty0
  (10):\penalty0 1331--1398.
\newblock \doi{https://doi.org/10.1002/cpa.21413}.
\newblock URL \url{https://onlinelibrary.wiley.com/doi/abs/10.1002/cpa.21413}.

\bibitem[Mallat(2016)]{Mallat2016}
S.~Mallat.
\newblock Understanding deep convolutional networks.
\newblock \emph{Philosophical Transactions of the Royal Society A:
  Mathematical, Physical and Engineering Sciences}, 374\penalty0
  (2065):\penalty0 20150203, 2016.
\newblock \doi{10.1098/rsta.2015.0203}.
\newblock URL
  \url{https://royalsocietypublishing.org/doi/abs/10.1098/rsta.2015.0203}.

\bibitem[Meyer et~al.(2024)Meyer, Dimitriadou, Hornik, Weingessel, and
  Leisch]{e1071}
D.~Meyer, E.~Dimitriadou, K.~Hornik, A.~Weingessel, and F.~Leisch.
\newblock \emph{e1071: Misc Functions of the Department of Statistics,
  Probability Theory Group (Formerly: E1071), TU Wien}, 2024.
\newblock URL \url{https://CRAN.R-project.org/package=e1071}.
\newblock R package version 1.7-16.

\bibitem[Oyallon and Mallat(2015)]{OyallonMallat2015}
E.~Oyallon and S.~Mallat.
\newblock Deep roto-translation scattering for object classification.
\newblock In \emph{2015 IEEE Conference on Computer Vision and Pattern
  Recognition (CVPR)}, pages 2865--2873, Los Alamitos, CA, USA, jun 2015. IEEE
  Computer Society.
\newblock \doi{10.1109/CVPR.2015.7298904}.
\newblock URL
  \url{https://doi.ieeecomputersociety.org/10.1109/CVPR.2015.7298904}.

\bibitem[Oyallon et~al.(2017)Oyallon, Belilovsky, and
  Zagoruyko]{OyallonETAL2017}
E.~Oyallon, E.~Belilovsky, and S.~Zagoruyko.
\newblock Scaling the scattering transform: Deep hybrid networks.
\newblock In \emph{2017 IEEE International Conference on Computer Vision
  (ICCV)}, pages 5619--5628, Los Alamitos, CA, USA, oct 2017. IEEE Computer
  Society.
\newblock \doi{10.1109/ICCV.2017.599}.
\newblock URL \url{https://doi.ieeecomputersociety.org/10.1109/ICCV.2017.599}.

\bibitem[Oyallon et~al.(2019)Oyallon, Zagoruyko, Huang, Komodakis,
  Lacoste-Julien, Blaschko, and Belilovsky]{OyallonETAL2019}
E.~Oyallon, S.~Zagoruyko, G.~Huang, N.~Komodakis, S.~Lacoste-Julien,
  M.~Blaschko, and E.~Belilovsky.
\newblock Scattering networks for hybrid representation learning.
\newblock \emph{IEEE Transactions on Pattern Analysis and Machine
  Intelligence}, 41\penalty0 (9):\penalty0 2208--2221, 2019.
\newblock \doi{10.1109/TPAMI.2018.2855738}.

\bibitem[Percival and Walden(2000)]{PercivalWalden2000}
D.~B. Percival and A.~T. Walden.
\newblock \emph{Wavelet Methods for Time Series Analysis}.
\newblock Cambridge Series in Statistical and Probabilistic Mathematics.
  Cambridge University Press, 2000.

\bibitem[{R Core Team}(2024)]{RSoftaware}
{R Core Team}.
\newblock \emph{R: A Language and Environment for Statistical Computing}.
\newblock R Foundation for Statistical Computing, Vienna, Austria, 2024.
\newblock URL \url{https://www.R-project.org/}.

\bibitem[Sifre and Mallat(2013)]{LaurentMallat2013}
L.~Sifre and S.~Mallat.
\newblock Rotation, scaling and deformation invariant scattering for texture
  discrimination.
\newblock In \emph{2013 IEEE Conference on Computer Vision and Pattern
  Recognition}, pages 1233--1240, 2013.
\newblock \doi{10.1109/CVPR.2013.163}.

\bibitem[Villar-Corrales and Morgenshtern(2021)]{VillarMorgenshtern2021}
A.~Villar-Corrales and V.~I. Morgenshtern.
\newblock Scattering transform based image clustering using projection onto
  orthogonal complement.
\newblock In \emph{Proceedings of the 2021 ACM Workshop on Intelligent
  Cross-Data Analysis and Retrieval}, ICDAR '21, page 24–32, New York, NY,
  USA, 2021. Association for Computing Machinery.
\newblock ISBN 9781450385299.
\newblock \doi{10.1145/3463944.3469098}.
\newblock URL \url{https://doi.org/10.1145/3463944.3469098}.

\bibitem[Whitcher(2024)]{waveslim}
B.~Whitcher.
\newblock \emph{waveslim: Basic Wavelet Routines for One-, Two-, and
  Three-Dimensional Signal Processing}, 2024.
\newblock URL \url{https://CRAN.R-project.org/package=waveslim}.
\newblock R package version 1.8.5.

\bibitem[Wickham et~al.(2019)Wickham, Averick, Bryan, Chang, McGowan,
  François, Grolemund, Hayes, Henry, Hester, Kuhn, Pedersen, Miller, Bache,
  Müller, Ooms, Robinson, Seidel, Spinu, Takahashi, Vaughan, Wilke, Woo, and
  Yutani]{tidyverse}
H.~Wickham, M.~Averick, J.~Bryan, W.~Chang, L.~D. McGowan, R.~François,
  G.~Grolemund, A.~Hayes, L.~Henry, J.~Hester, M.~Kuhn, T.~L. Pedersen,
  E.~Miller, S.~M. Bache, K.~Müller, J.~Ooms, D.~Robinson, D.~P. Seidel,
  V.~Spinu, K.~Takahashi, D.~Vaughan, C.~Wilke, K.~Woo, and H.~Yutani.
\newblock Welcome to the {tidyverse}.
\newblock \emph{Journal of Open Source Software}, 4\penalty0 (43):\penalty0
  1686, 2019.
\newblock \doi{10.21105/joss.01686}.

\bibitem[Zarka et~al.(2020)Zarka, Thiry, Angles, and Mallat]{zarkaETAL2020}
J.~Zarka, L.~Thiry, T.~Angles, and S.~Mallat.
\newblock Deep network classification by scattering and homotopy dictionary
  learning, 2020.

\bibitem[Zarka et~al.(2021)Zarka, Guth, and Mallat]{ZarkaETAL2021separation}
J.~Zarka, F.~Guth, and S.~Mallat.
\newblock Separation and concentration in deep networks, 2021.

\end{thebibliography}

\end{document}